\newif\ifarxiv
\newif\ifiros
\newif\ifirosfinal

\irosfalse \irosfinalfalse \arxivtrue %

\documentclass[letterpaper, 10 pt, conference]{ieeeconf}  %

\IEEEoverridecommandlockouts                              %

\usepackage{graphics} %
\usepackage{epsfig} %
\usepackage{mathptmx} %
\usepackage{times} %
\usepackage{amsmath} %
\usepackage{amssymb}  %

\usepackage{cite}
\usepackage{multirow}
\usepackage{arydshln} %

\usepackage[dvipsnames]{xcolor}
\usepackage{xcolor,colortbl}
\usepackage{caption}
\makeatletter
\let\NAT@parse\undefined
\makeatother
\definecolor{citepurple}{rgb}{0.288,0.1196,0.7}
\definecolor{darkgreen}{rgb}{0.0, 0.5, 0.0}
\definecolor{darkblue}{rgb}{0.0, 0.0, 0.7}
\usepackage[pagebackref,breaklinks,colorlinks,bookmarks,citecolor=citepurple]{hyperref}

\usepackage{booktabs, multirow}

\definecolor{Gray}{gray}{0.50}
\newcolumntype{g}{>{\columncolor{Gray}}c}
\definecolor{ffe1da}{RGB}{255,225,218}
\definecolor{F7E0D5}{RGB}{247,224,213}
\definecolor{darkF7E0D5}{RGB}{209,154,128}
\colorlet{Light}{White!0!F7E0D5}

\colorlet{tabfirst}{Green!25}
\definecolor{tabthird}{rgb}{1, 0.85, 0.7}
\definecolor{tabsecond}{rgb}{1, 0.96, 0.7}

\definecolor{darkorange}{rgb}{1.0, 0.54, 0}
\definecolor{darkpurple}{rgb}{0.288,0.1196,0.7}
\definecolor{amber}{rgb}{1.0, 0.75, 0.0}

\usepackage[capitalize]{cleveref}
\crefname{section}{Section}{Sections}
\crefname{table}{Table}{Tables}

\newcommand{\xxnote}[3]{}
\ifx\hidenotes\undefined
  \renewcommand{\xxnote}[3]{\color{#2}{#1: #3}}
\fi

\newcommand{\authorhref}[3][ourpurple]{\href{#2}{\color{#1}{#3}}}

\newcommand{\mapname}{\textbf{\textbf{\texttt{RayFronts}}}}

\newcommand{\webpage}{rayfronts.github.io}
\definecolor{ourorange}{rgb}{0.9, 0.34, 0.145}
\definecolor{ourpurple}{rgb}{0.2431, 0.26, 0.6}
\definecolor{ourpink}{rgb}{0.835,0.1019, 0.4078}

\ifarxiv
\title{\LARGE \bf
RayFronts: Open-Set Semantic Ray Frontiers for \\
Online Scene Understanding and Exploration \\[2 pt]
\Large{\href{https://\webpage/}{\color{ourorange}{\webpage}}}
\vspace{-1em}
}
\fi
\ifiros
\title{\LARGE \bf
RayFronts: Open-Set Semantic Ray Frontiers for\\Online Scene Understanding and Exploration}
\fi
\ifirosfinal
\title{\LARGE \bf
RayFronts: Open-Set Semantic Ray Frontiers for\\Online Scene Understanding and Exploration}
\fi

\ifarxiv
\author{
\authorhref{https://www.linkedin.com/in/omaralama/}{Omar Alama},
\authorhref{https://www.linkedin.com/in/avigyan-bhattacharya}{Avigyan Bhattacharya}, 
\authorhref{https://purenothingness24.github.io/}{Haoyang He}, 
\authorhref{https://seungchan-kim.github.io/}{Seungchan Kim}, 
\authorhref{https://haleqiu.github.io/}{Yuheng Qiu}, 
\\ 
\authorhref{https://theairlab.org/team/wenshan/}{Wenshan Wang}, 
\authorhref{https://cherieho.com/}{Cherie Ho}, 
\authorhref{https://nik-v9.github.io/}{Nikhil Keetha}, 
\authorhref{https://theairlab.org/team/sebastian/}{Sebastian Scherer}
\\[5 pt]
\href{https://www.ri.cmu.edu/}{\color{ourpink}{Carnegie Mellon University}}
\thanks{Authors are with Carnegie Mellon University, Pittsburgh, PA, USA. \tt{Emails: \{oalama, avigyanb, hhe2, seungch2, yuhengq, wenshanw, cherieh, nkeetha, basti\}@andrew.cmu.edu}
}%
}
\fi
\ifiros
\author{
Omar Alama, 
Avigyan Bhattacharya, 
Haoyang He, 
Seungchan Kim, 
Yuheng Qiu, 
\\ 
Wenshan Wang, 
Cherie Ho, 
Nikhil Keetha, 
Sebastian Scherer 
\thanks{This work was supported by DSTA Contract #DST000EC124000205, KAU, and NIDILRR
Grant #90IFDV0042}%
\thanks{Authors are with Carnegie Mellon University, Pittsburgh, PA, USA. \tt{Emails: \{oalama, avigyanb, hhe2, seungch2, yuhengq, wenshanw, cherieh, nkeetha, basti\}@andrew.cmu.edu}
}%
}
\fi

\begin{document}

\thispagestyle{empty}
\pagestyle{empty}

\makeatletter
\let\@oldmaketitle\@maketitle
\renewcommand{\@maketitle}{\@oldmaketitle
\centering
\captionsetup{type=figure, singlelinecheck=false}
\begin{tabular}{cccc}
\ifiros
\includegraphics[width=.99\textwidth]{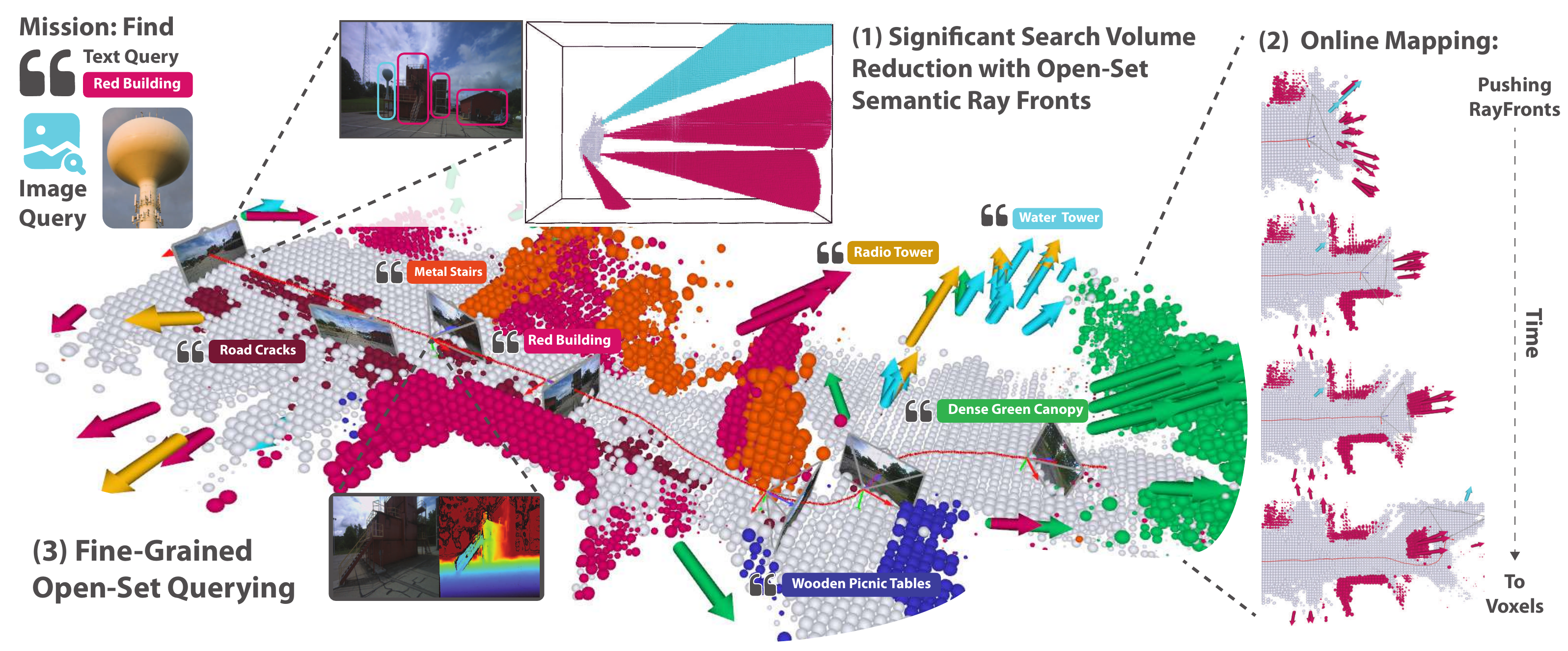}
\fi
\ifarxiv
\includegraphics[width=.99\textwidth]{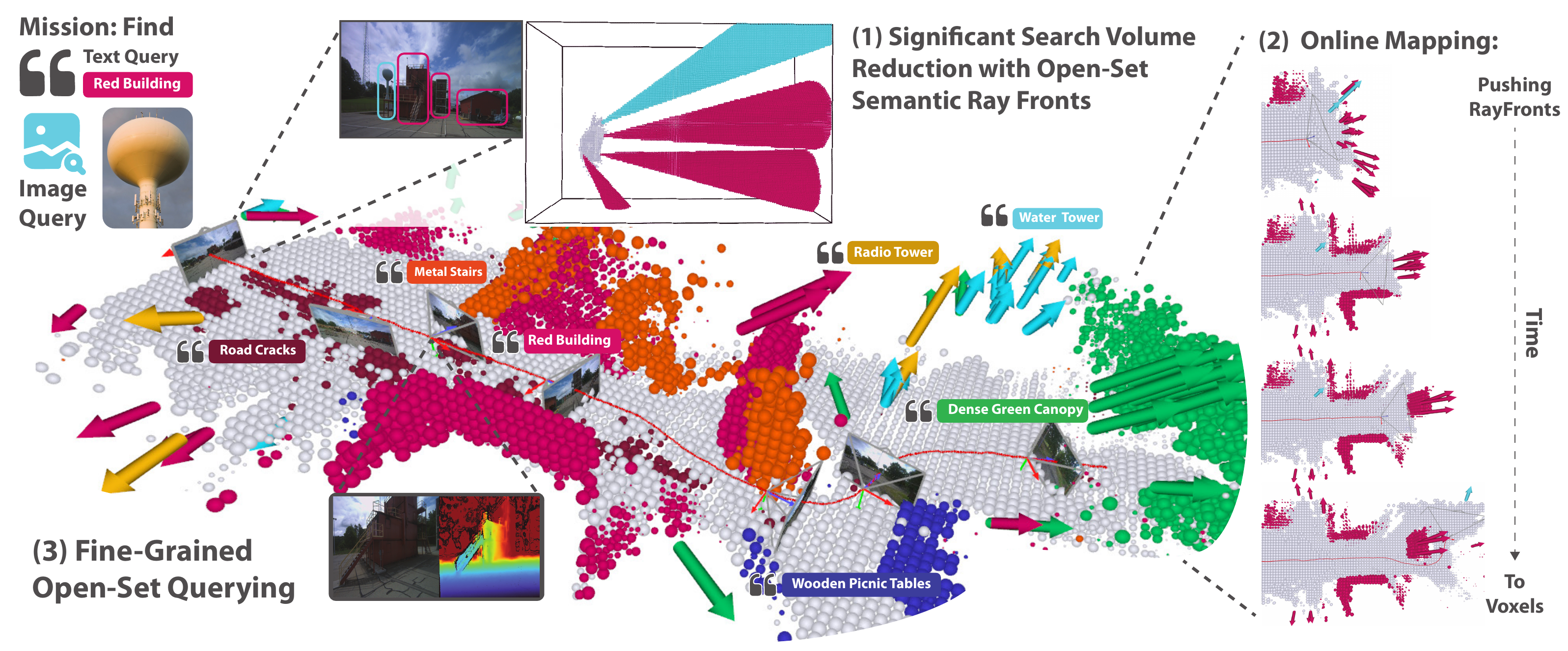}
\fi
\end{tabular}

\caption{\small \textbf{\mapname{} is a real-time semantic mapping system that enables fine-grained scene understanding both within and beyond the depth perception range.} Given an example mission through multi-modal queries to locate red buildings \& a water tower, \mapname{} enables: (1) Significant search volume reduction for online exploration (as shown by the red and blue cones at the top) and localization of far-away entities (e.g., the water \& radio tower). (2) Online semantic mapping, where prior semantic ray frontiers evolve into semantic voxels as entities enter the depth perception range (e.g., the red buildings query on the right side). (3) Multi-objective fine-grained open-set querying supporting various open-set prompts such as ``Road Cracks", ``Metal Stairs", and ``Green Dense Canopy".}
\label{fig:fig1}
\setcounter{figure}{1}
}
\makeatother

\maketitle

\begin{abstract}

Open-set semantic mapping is crucial for open-world robots.
Current mapping approaches either are limited by the depth range or only map beyond-range entities in constrained settings, where overall they fail to combine within-range and beyond-range observations.
Furthermore, these methods make a trade-off between fine-grained semantics and efficiency.
We introduce \mapname{}, a unified representation that enables both dense and beyond-range efficient semantic mapping.
\mapname{} encodes task-agnostic open-set semantics to both in-range voxels and beyond-range rays encoded at map boundaries, empowering the robot to reduce search volumes significantly and make informed decisions both within \& beyond sensory range, while running at $8.84$ Hz on an Orin AGX.
Benchmarking the within-range semantics shows that \mapname{}'s fine-grained image encoding provides \textbf{$1.34 \times$} zero-shot 3D semantic segmentation performance while improving throughput by \textbf{$16.5 \times$}.
Traditionally, online mapping performance is entangled with other system components, complicating evaluation.
We propose a planner-agnostic evaluation framework that captures the utility for online beyond-range search and exploration, and show \mapname{} reduces search volume \textbf{$2.2 \times$} more efficiently than the closest online baselines.

\end{abstract}

\section{Introduction}
\label{sec:introduction}

Open-set semantic mapping is essential for robotic systems to reason, search, and navigate in open-world environments. The task requires capturing both fine-grained local details and distant beyond-range semantic cues in real-time. For instance, as shown in \cref{fig:fig1}, an aerial or ground robot may need to localize the water or radio towers over 100 meters beyond its depth perception capability, as well as locate any hazards (road cracks) or interesting structures along the way (red building). This work explores what the most effective semantic mapping system would be to capture this information to imbue the robot with the ability to reason within and beyond depth sensing limitations.

Although there is a growing body of literature on open-set metric semantic mapping \cite{chen2023open, conceptfusion, hovsg-werby, conceptgraphs, peng2023openscene}, these methods focus primarily on offline mapping for downstream usage in limited environments ignoring efficiency and depth-sensing limitations. Such representations cannot guide the robot in search and exploration tasks as they provide no information about the unmapped region. Other works change the way semantics are typically encoded in a map (point clouds, voxels, and bounding boxes) to representations that can guide exploration (i.e semantic frontiers \cite{yokoyama2024vlfm, chen2023train}, and semantic poses \cite{thomas2024embedding}). However, existing semantic frontier maps are limited to 2D indoor environments and have limited semantics due to whole-image encoding \cite{yokoyama2024vlfm} or using closed-set models \cite{chen2023train}, whereas semantic poses \cite{thomas2024embedding} lack fine-grained reconstructions and can only recognize prominent objects in an image.

In this context, we explore the question, \textbf{``How to design an efficient online mapping representation that facilitates fine-grained scene understanding, and be aware of beyond-range semantic entities?''} 
We introduce \textbf{\mapname{}}, a semantic map representation which seamlessly integrates traditional within-depth mapping with ray-based representations, facilitating both dense mapping within observed depth ranges and perception beyond them.
Unlike conventional representations truncated at the depth range, multi-directional semantic ray frontiers retain coarse-grained far-range information, enabling downstream planners (e.g., object-search) to reduce their search volume significantly. Additionally, to assess the utility of the proposed representation, we construct a planner-agnostic benchmark and propose a new metric to measure how effectively an online mapping strategy reduces the search space for fast object localization and exploration. Finally, to avoid image level encoding and expensive pipelines, we introduce a novel image encoder that achieves state-of-the-art performance on zero-shot 3D semantic segmentation enabling a computationally efficient, open-world, and deployable 3D online mapping system. 

Our key contributions are as follows: \\
\textbf{C1: Unified 3D Map Representation for Within-Depth and Beyond-Depth Perception:} We develop the first-of-its-kind open-set semantic ray frontier 3D map, which enables robots to reason in open environments achieving up to 1.85x mIoU in offline zero-shot performance, and are 2.2x more efficient in reducing search volume in online mapping than the closest offline \& online baselines respectively.

\noindent\textbf{C2: Planner-Agnostic Online Semantic Mapping Evaluation Framework:} We showcase that online semantic mapping systems can be evaluated on their fundamental utility for exploration, without being tightly coupled with a planner, by developing a metric that assesses ``correctly reduced search volume".

\noindent\textbf{C3: Efficient real-time open-set online mapping system:} can run end to end at 8.84 Hz on an ORIN AGX and our efficient dense vision-language encoder is 16.5x faster than the closest baseline and achieves state-of-the-art on open-vocab zero-shot 3D semantic segmentation mIoU.

\section{Related Work}
\label{sec:related-work}

\subsection{Dense 2D Open-Set Semantics}

The rapid rise of foundation models~\cite{bommasani2021opportunities} has spearheaded progress in tasks requiring fine-grained open-set concepts which are hard to capture with a fixed taxonomy of semantic classes~\cite{oquab2023dinov2, SAM}.
CLIP~\cite{radford2021learning} and its subsequent variants such as SIGLIP~\cite{zhai2023sigmoid} have shown impressive alignment between abstract textual concepts and images.
These Visual Language Models (VLMs) initially aligned textual descriptions and images as a whole and not to particular regions or pixels.
Subsequently, follow-up work based on supervised and unsupervised regimes has attempted to address this issue~\cite{wu2024towards}.
A recurring theme in these methods is the trade-off between efficiency and accuracy, with the most performant approaches often using multiple foundation models like DINOv2~\cite{oquab2023dinov2}, Grounding DINO~\cite{liu2024grounding}, and SAM~\cite{SAM}.
This is not optimal for online real-world deployment and hence we explore the applicability of RADIO~\cite{RADIO}, a foundation model aligned with various dense visual foundation models.
While RADIO's language alignment is to the image as a whole, we find that employing a simple attention trick~\cite{naclip} with its SIGLIP adapter enables us to achieve state-of-the-art pixel-level language alignment and real-time performance on embedded hardware.

\subsection{Offline \& Bounded Open-Set Semantic Mapping}

Traditional semantic mapping systems have relied on learning-based methods to detect and segment a fixed set of concepts, with performance limited by vocabulary size and training distribution~\cite{semanticfusion, runz2018maskfusion, grinvald2019volumetric, fusion++, mid-fusion, hughes2022hydra, ho2024map}. With the rise of dense 2D open-set semantics, interest has grown in open-vocab semantic mapping systems using representations like point clouds, voxels, and scene graphs~\cite{hu2023toward}. These systems have shown strong open-world capabilities for navigation, manipulation, and scene understanding~\cite{chen2023open, conceptfusion, hovsg-werby, conceptgraphs, peng2023openscene, keetha2023anyloc, xie2024embodied, kassab2024language}. However, most focus on offline database maps and lack online utility for robotics, with many design choices making real-time deployment infeasible. To address this, we introduce a computationally efficient, fast, and deployable 3D mapping system for online scene understanding.

\begin{figure*}[ht]
    \centering
    \ifiros
    \includegraphics[width=\textwidth]{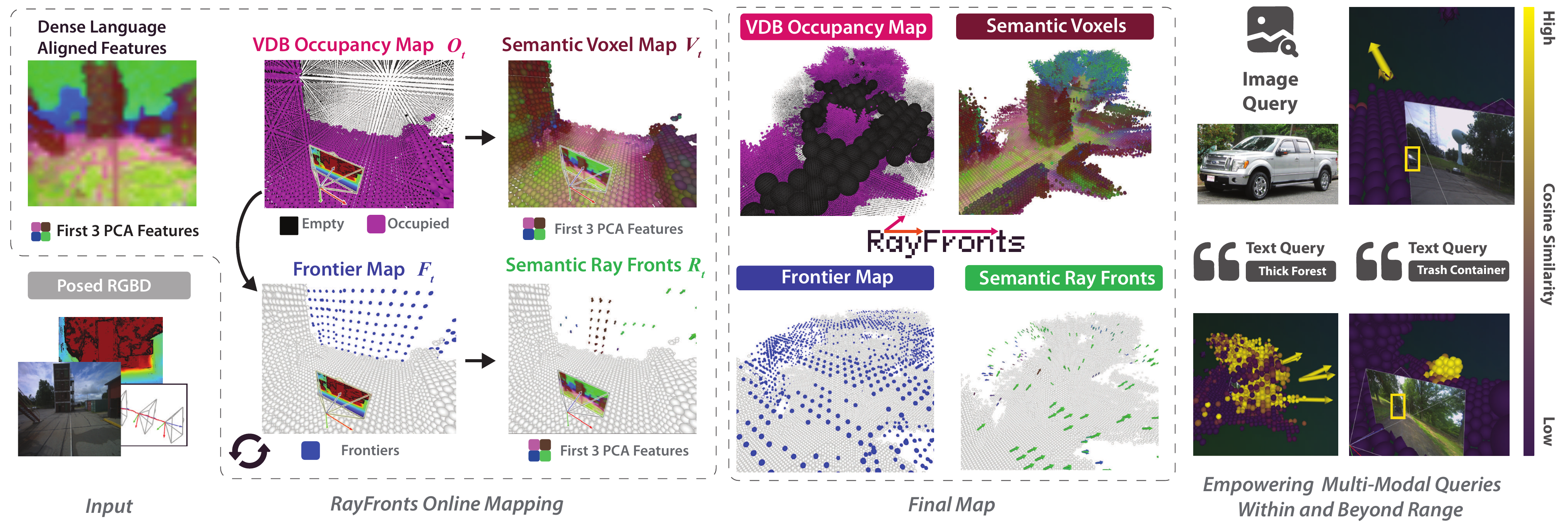}
    \fi
    \ifarxiv
    \includegraphics[width=\textwidth]{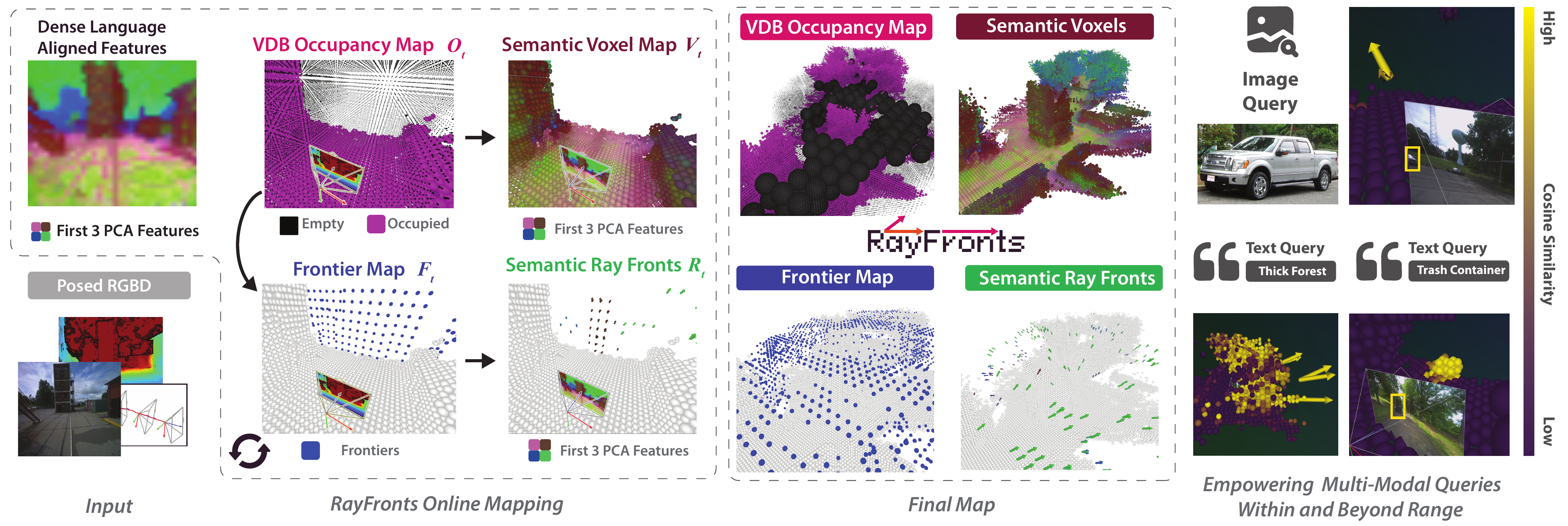}
    \fi
    \caption{\small \textbf{Overview of our online mapping system, \mapname{} is designed for multi-objective \& multi-modal open-set querying of both in-range and beyond-range semantic entities.} Given posed RGB-D images, we first extract dense features with our fast language-aligned image encoder. Then, posed depth information and features are used to construct a semantic voxel map for in-range queries. In parallel, \mapname{} also maintains a VDB-based occupancy map to generate frontiers, which are further associated with multi-directional semantic rays. These semantic ray fronts enable us to perform beyond-range querying of open-set concepts in the unobserved region.}
    \label{fig:system}
\end{figure*}

\subsection{Online \& Unbounded Open-Set Semantic Mapping}

While offline and bounded semantic mapping has excelled in indoor scenes, it struggles with outdoor, unbounded, and unstructured environments, where limited depth perception becomes a challenge. An effective online semantic mapping system must support both efficient exploration and fine-grained scene understanding. VLFM~\cite{yokoyama2024vlfm} addresses this by encoding semantics on 2D frontiers for object goal navigation, but it is limited to a single object at a time and only works in indoor settings. Similarly, Embedding Pose Graph (EPG)~\cite{thomas2024embedding} encodes semantics into rays from pose nodes, but lacks fine-grained mapping and condenses the entire image into one feature vector, risking the loss of subtle details.

In contrast, we propose a novel representation combining semantic voxels with ray-based frontiers, capturing multiple viewing directions and open-set features. This approach enables efficient online search and rough triangulation of distant objects, allowing us to capture both in-range and beyond-range semantic entities. Our synergy of metric-map-based semantic voxels and direction-based ray frontiers supports fine-grained scene understanding and efficient exploration.

\section{Method}
\label{sec:method}

We present \mapname{}, a unified 3D semantic mapping system for multi-modal open-set semantic querying of both in-range and beyond-range semantic entities. \mapname{} maintains a semantic voxel map $\mathcal{V}_t$ containing voxel coordinates and semantic features for within-range entities, an occupancy VDB map $\mathcal{O}_t$, a set of frontiers $\mathcal{F}_t$ denoting subsampled boundary voxels between observed and unobserved spaces, and semantic ray fronts $\mathcal{R}_t$, a ray-based representation on the frontiers, which contains features for beyond-range semantic reasoning. 

\mapname{} operates in four steps: (1) extracting dense, language-aligned features from RGB input through our efficient encoding pipeline, (2) fusing within-range featurized points into a sparse semantic voxel map, (3) maintaining an occupancy map for frontier computation and semantic voxel pruning, and (4) ray casting beyond-range semantics onto frontiers to semantically reason beyond the observed map. Our system is optimized for parallel computing and online mapping, leveraging PyTorch tensors for $\mathcal{V}_t, \mathcal{F}_t, \mathcal{R}_t$ on the GPU and OpenVDB \cite{museth2013openvdb} for $\mathcal{O}_t$ on the CPU. This design ensures efficient querying, seamless feature integration, and adaptability to evolving environments. The pipeline and outputs of \mapname{} are illustrated in \cref{fig:system}.

\subsection{Extracting Dense Language-Aligned Features}
\label{sec:method-dense-features}
There has been a rapid growth of methods that extract dense language aligned features from RGB images. However, existing methods fall short by (1) lacking generalization due to limited supervision, (2) sacrificing efficiency with multi-model multi-stage pipelines, or (3) prioritizing efficiency and generalization at the cost of segmentation quality. In this work,  we adopt RADIO\cite{RADIO}, a VFM that distills key features from CLIP\cite{radford2021learning}, DINOv2\cite{oquab2023dinov2}, and SAM\cite{SAM}. This integration enhances feature representation and segmentation performance. 
However, since RADIO leverages vanilla ViT \cite{dosovitskiy2020image}, it struggles with fine-grained localization of visual features, a critical challenge in semantic scene understanding. To address this, we integrate the explicit spatial attention mechanism proposed by NACLIP \cite{naclip} and modify the RADIO encoder accordingly. 
Specifically, we augment the attention layer of the final ViT block by introducing a locality constraint via an unnormalized multivariate Gaussian kernel centered around each patch essentially pushing the model to attend to its neighboring patches and improving locality.

To densely align the RADIO feature space with language, we explore the available pre-trained MLP-based adaptor heads provided by RADIO. Simply following RADIO's original distillation approach-- projecting spatial features onto CLIP or SIGLIP space using their respective adapters--yields subpar performance. Instead, we use the SIGLIP summary feature adapter to project spatial features to the SIGLIP \texttt{CLS} token space, thus resulting in a spatially consistent and language-aligned feature map and observe significant performance improvements over existing methods.

\subsection{Semantic Voxels for Dense Within-Depth Queries}
\label{sec:method-sem-voxels-within-depth}
Given a pose $P_t \in \mathbb{SE}3$ and depth map $D_t \in \mathbb{R}^{H\times W}$, we initialize a voxel grid retaining only points within the frustrum, yielding $Q_t$.
We transform the points $Q_t$ into the camera frame and classify its occupancy based on depth $D_t$. For each occupied point, we find the associated feature via nearest-neighbor interpolation yielding
$\mathcal{P}_t^{local} = \{ (\mathbf{p}_i, \mathbf{f}_i) \}_{i=1}^{M}$ where $\mathbf{p}_i \in \mathbb{R}^3$ are point coordinates and $\mathbf{f}_i \in \mathbb{R}^{3+D+1}$ (3 for RGB, $D$ is feature dimension, and 1 for the hit count). Local updates are accumulated into a buffer of $m$ frames before being voxelized at resolution $\alpha$ and integrated into the global map $\mathcal{V}$.

\textbf{Feature Fusion and Aggregation}: Rather than complex fusion methods used in \cite{conceptfusion, hovsg-werby}, we employ a simple weighted average, where each voxel's hit count serves as the weight when fusing features within the same voxel. To achieve this, we concatenate coordinate and feature tensors of accumulated local updates $\mathcal{P}_t^{local}$ with those of global voxel map $\mathcal{V}_t$. 
A parallel scatter-reduce operation fuses features at the same discretized coordinates into a single voxel.

\subsection{Occupancy Mapping for Frontiers and Pruning}
\label{sec:method-occupancy}
To represent occupancy map $\mathcal{O}_t$ efficiently, we employ OpenVDB \cite{museth2013openvdb}, recently used in modern 3D frontier-based exploration works \cite{best2022resilient, hagmanns2022efficientglobaloccupancymapping, kim2023multi} for its sparse tree representation and multi-resolution capability. Following standard practice, we store log-odds occupancy $o_j$ in a signed byte.
To better tolerate dynamic environments and to avoid overflow, we limit $prob_{occ}(o_j)$ to lower and upper limits. \cref{fig:system} shows the OpenVDB map with large free voxels showing the multi-resolution aspect of the occupancy representation.

\textbf{Pruning Semantic Voxels:} When accumulating voxels over long distances and time periods, odometry drifts and dynamic objects can introduce inconsistencies, not to mention the growing memory consumption. To mitigate this, we prune invalid semantic voxels by querying the occupancy map $\mathcal{O}_t$ and removing those with occupancy below 0.5.

\subsection{Finding the ``Fronts": Computing 3D Frontiers}
\label{sec:method-finding-fronts}
We identify frontiers by iterating over all free observed voxels using efficient OpenVDB iterators and examining their neighbors. A voxel is considered a frontier if its neighbors meet the minimum thresholds for unobserved ($min_{unobsrv}$), occupied ($min_{occ}$), and free ($min_{free}$) counts, allowing us to emphasize frontiers near surfaces or open space as needed. 
To reduce density, we subsample the frontier map using a coarser voxel grid of size $\beta$.
Fine-grid frontiers are accumulated into a coarser grid, and cells with enough frontiers remain as frontiers. 

\subsection{Semantic Ray Frontiers for Beyond-Depth Mapping}
\label{sec:method-semantic-ray-fronts}
\textbf{Need for richer frontiers: } Existing semantic frontier methods have fundamentally constrained beyond-range semantic encoding, where only a single object can be pursued at a time due to feature collisions from distinct objects observed through the same frontier.
To enable multi-object semantic guidance for search and exploration, we transition from conventional semantic frontiers $\mathcal{F}_{sem} = \{(\mathbf{p}_k, \mathbf{f}_k)\}_{k=1}^{F}$
to semantic ray frontiers 
$\mathcal{R}_{sem}= \{ (\mathbf{o}_r,\mathbf{\theta}_r, \mathbf{\phi}_r, \mathbf{f}_r) \}_{r=1}^{R}$ 
, where $\mathbf{o}_r$ is ray origin, $\mathbf{\theta}_r \in [-\pi, \pi)$ and $\mathbf{\phi}_r \in [0, \pi)$ are azimuthal and zenith angles, and $\mathbf{f}_r$ are semantic features. This shift drastically enhances the mapping system by allowing efficient storage of rich multi-object semantics with minimal feature collisions, enabling rough triangulation of object locations, and reducing the search space volume needed for exploration. We discuss the ray mapping process (observe, associate, discretize \& accumulate) and how rays are pruned and propagated below.

\textbf{Observe: } To identify out-of-range regions in the feature map $F_t$ we compute a boolean mask $M_t \in \mathbb{R}^{H\times W}$ from the depth $D_t$ (obtained via stereo, LiDAR, or monocular depth estimation). The mask encompasses either $+\infty$ values from depth sensors or far low-certainty values. $M_t$ is eroded to prevent semantic leakage at object boundaries, and used to select the semantic pixels to propagate as rays $\mathcal{R}_t^{local} = \{(\mathbf{o}_r,\mathbf{d}_r, \mathbf{f}_r)\}_{r=0}^{H_t}$ where  \( \mathbf{o}_r \in \mathbb{R}^3 \) is the ray and camera origin, \( \mathbf{d}_r \in \mathbb{R}^3 \) is the normalized direction vector, and $\mathbf{f}_r$ represents semantic features. 

\textbf{Associate (Matching Rays to Frontiers):} In the presence of depth information, rather than keeping rays at the robot's origin as in \cite{thomas2024embedding}, we leverage the mapped area to push rays closer to their observed entities, improving localization. For each semantic ray $(\mathbf{o}_r,\mathbf{d}_r, \mathbf{f}_r)$, we select a frontier from the candidate set $\mathcal{F}_{t+1}$ through a two-step filtering process. First, we prune frontiers  by (1) removing those not in front of the ray (2) computing the shortest orthogonal distance $d_{ortho}$ between the ray and frontiers, discarding those where $d_{ortho} > \beta$ (exceeding the frontier grid cell size), and (3) calculating the distance from ray origin $\mathbf{o}_r$ to frontier origin $\mathbf{p}$, obtaining $d_{orig}$ and removing frontiers where $d_{orig} > 4 \times \textit{depth}\_\textit{range}$. 

Next, for the remaining $k$ candidate frontiers, we compute a cost function \begin{equation}
    d_{cost} = (\frac{d_{ortho}}{max(\{d_{ortho}^r\}_{r=0}^{k})} + \frac{d_{orig}}{max(\{d_{orig}^r\}_{r=0}^{k})}) / 2 , d_{cost} \in [0, 1]
\end{equation}We select the frontier with the minimum $d_{cost}$ as the best match. We qualitatively find that utilizing both $d_{ortho}$ and $d_{orig}$ improves results and prevents distant frontiers from receiving noisy semantics. 

For further refinement, we optionally apply ray tracing, marching each ray through the occupancy map $\mathcal{O}_{t+1}$ until it reaches its assigned frontier or encounters occupied or unobserved (possibly occupied) cells. At this stage, each semantic ray is associated with a frontier, updating its origin ($\mathbf{o}_r$) to the corresponding frontier origin $\mathbf{p}$. Since we lack depth information about the underlying semantic entity, we maintain the ray's direction $\mathbf{d}_r$ when shifting its origin.

\textbf{Discretize and Accumulate (``Ray Binning"):} Similar to voxelization techniques, we organize semantic rays into angle bins with a resolution of $\psi$ degrees. The normalized ray directions $\mathbf{d}_r$ are converted to spherical angles using:
$\mathbf{\theta_r} = atan2(d_r^1, d_r^0), \quad \mathbf{\phi_r} = acos(d_r^2)$
where $atan2$ is the four-quadrant inverse tangent. We then discretize these angles and merge rays that correspond to the same frontier and the same angle bin from both the local update $\mathcal{R}_t^{local}$ and the global set $\mathcal{R}_{t}$. We use $1-d_{cost}$ for weighing the features while merging, assigning lower trust to high-cost associations. This yields the updated ray-frontier $\mathcal{R}_{t+1}$.

\textbf{Pushing the ray fronts onward:} Semantic ray frontiers must be updated as new areas are mapped. After computing the frontier update $\mathcal{F}_{t+1}$, we use a voxel grid to perform a set intersection between all ray origins and frontier origins, similar to pruning voxels, and remove rays no longer associated with active frontiers.
If ray tracing is enabled, removed rays are added back to the ray accumulation buffer to be re-cast in the next iteration. This preserves previously observed semantics that may no longer be directly visible as the frontier shifts (e.g., from a side view). However, without ray tracing, removed rays would continue to propagate indefinitely, so we disable the behavior under that setting. In our experiments, we use ray tracing unless otherwise stated.

\begin{figure}[!tb]
\centering
\includegraphics[width=0.7\linewidth]{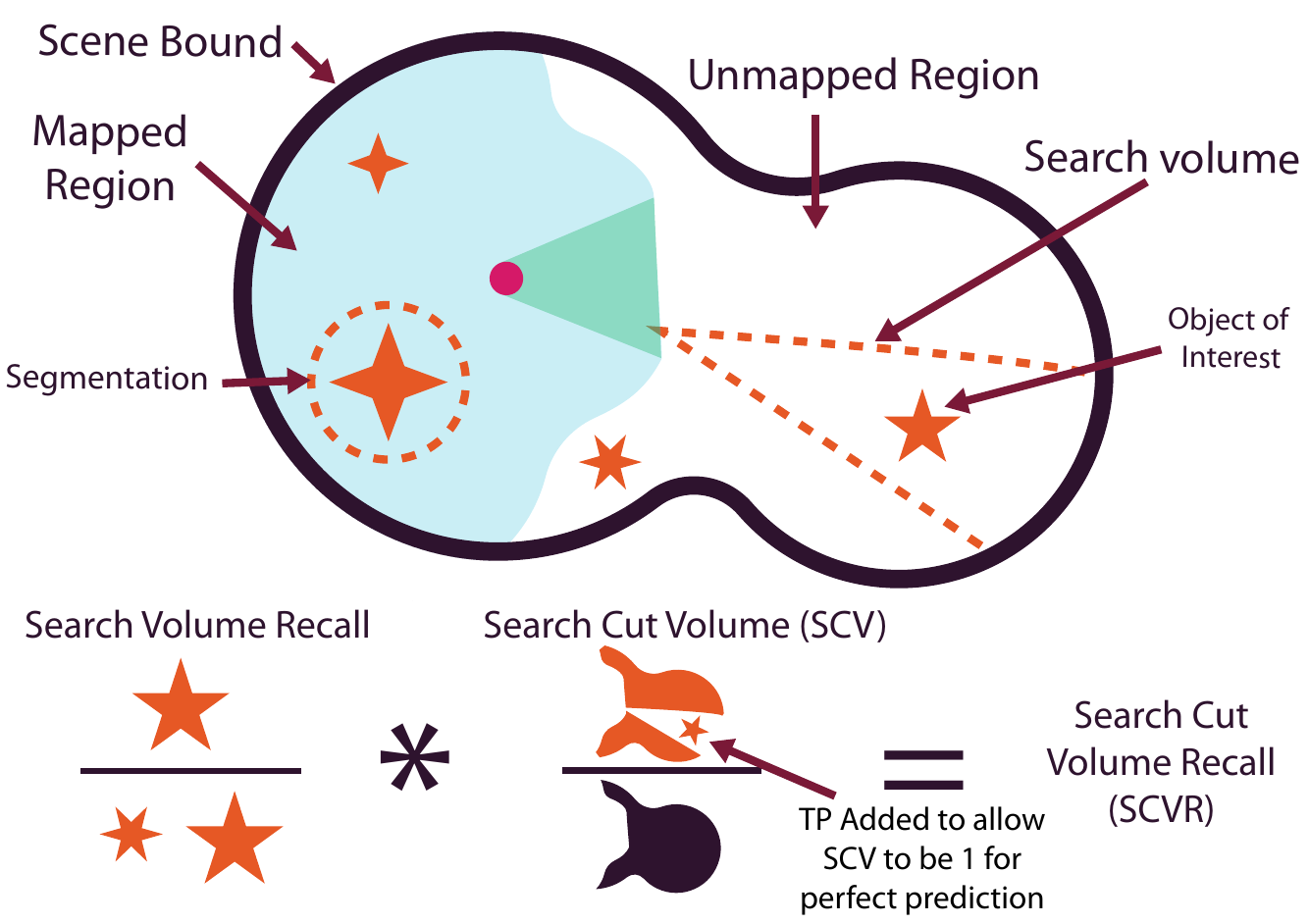} 
\caption{\textbf{An illustration of our proposed planner-agnostic metric (Search Cut Volume Recall) for open-world online search benchmarking.} Intuitively, the metric captures ``How much of the search volume is eliminated correctly?" An optimal mapper should promptly and accurately reduce the search space, enabling fast multi-object localization and exploration.}
\label{fig:metrics}
\end{figure}

\section{Experimental Setup}
\label{sec:setup}
A good online mapping system should (1) intelligently guide the robot toward regions of interest in any environment, eliminating irrelevant volumes early, (2) accurately capture fine-grained open-set semantics within a metric map, and (3) do so efficiently. In this section, we first introduce our proposed online mapping evaluation framework, which assesses the utility of a mapping system in guiding exploration without a planner in-the-loop, and introduce competitive baseline representations. We then present our extensive offline map evaluation following established protocols. Finally, we conduct a deployability and throughput analysis.

\subsection{Planner-Agnostic Online Semantic Mapping Evaluation}
\label{sec:experiments-online-mapping}
\noindent\textbf{Dataset:} Originally designed to challenge visual SLAM with large, cluttered, long-tail objects in indoor and outdoor environments, TartanAirV2\cite{wang2020tartanair} serves as a stress test for our representation. To simulate scenarios with severely limited depth, we choose four large outdoor scenes \texttt{AbandonedCableday, Factory, Downtown} and \texttt{ConstructionSiteOvercast} where bounding boxes span approximately 8 million $m^3$ with a 50m range cutoff. We generate ground truth occupancy (defining the scene volume) and a semantic label map at 1-meter voxel resolution from the provided posed RGBD input.

\noindent\textbf{Baselines:}
There are no established online mapping baselines for 3D open-world environments. Therefore, we take inspiration from existing works and design the following baselines, keeping the encoder fixed to isolate the impact of our mapping approach:
\begin{itemize}
    \item \textbf{Semantic Poses} (Sem Pose): Emulates EPG \cite{thomas2024embedding} by using global encoding for the image, resulting in a single ray per frame located at the robot origin.
    \item \textbf{Semantic Voxels} (Sem Voxels): Subsumes representations that encode only within-range semantics \cite{hovsg-werby, conceptfusion, conceptgraphs}.
    \item \textbf{Semantic Frontiers}: Emulates in 3D the 2D approaches that paint frontiers with semantics \cite{yokoyama2024vlfm, chen2023train}. We recognize that there are two ways semantic frontiers can be interpreted; (1) As a spherical region encompassing the semantic entity (i.e \textbf{Spherical Sem Fronts}), or (2) as a single ray pointing away from the observed region (i.e \textbf{Unidirectional Sem Fronts}). We evaluate both. 
\end{itemize}
We define search volume as the unmapped region unless further evidence is provided. Ray-based approaches cast search cones while spherical sem fronts define a sphere volume extending to the nearest frontier. Multiple search volumes are summed in a voxel grid, counts are normalized, and thresholded at 0.05 to get the final search volume for a class. For \textbf{Unidirectional Sem Fronts}, frontier directions are inferred using the occupancy map $\mathcal{O}_t$ by computing a weighted combination of all directions around a frontier in a 3x3x3 window where mapped voxels have a weight of -1 (Pushing away) and unmapped voxels have a weight of +1 (Pulling toward).

\noindent\textbf{Evaluation Protocol:}
We ask the question ``Can an online semantic mapping system's utility for search and exploration be assessed independently of specific planners ?" Yes, the key is examining how accurately and efficiently the map constrains the search space. Traditional mapping metrics such as mIoU, mAcc, F1 measure fine-grained semantic localization but overlook search volume efficiency, as they ignore \textbf{true negatives}. In search and exploration, a high true negative rate in the unobserved region reduces wasted search time. Therefore, for beyond-range search volume estimation, we introduce a novel metric below, and to evaluate within-range fine-grained online performance, we use the area under the mIoU-time curve.
\begin{table*}[!t]\centering
\caption{Online \& Unbounded Semantic Mapping Benchmarking on TartanAirV2~\cite{wang2020tartanair}. Ranking shown as \colorbox{green!25}{\textbf{first}}, \colorbox{yellow!30}{second}, and \colorbox{orange!30}{third}.}\label{tab:mapping}
\scriptsize
\resizebox{\textwidth}{!}{
\begin{tabular}{lccccccccccccc}\toprule
&\multicolumn{4}{c}{\textbf{0m Depth (AUC)}} &\multicolumn{4}{c}{\textbf{10m Depth (AUC)}} &\multicolumn{4}{c}{\textbf{20m Depth (AUC)}} \\
\cmidrule{2-5} \cmidrule(lr{0.75em}){6-9} \cmidrule(lr{0.75em}){10-13}
\textbf{Methods} &mIoU(\%) &SCV(\%) &Recall(\%) &SCVR(\%) &mIoU(\%)  &SCV(\%) &Recall(\%) &SCVR(\%) &mIoU(\%) &SCV(\%)  &Recall(\%) &SCVR(\%)  \\
\cmidrule{1-1} \cmidrule(lr{0.75em}){2-2} \cmidrule(lr{0.75em}){3-3} \cmidrule(lr{0.75em}){4-4} \cmidrule(lr{0.75em}){5-5} \cmidrule(lr{0.75em}){6-6} \cmidrule(lr{0.75em}){7-7} \cmidrule(lr{0.75em}){8-8} \cmidrule(lr{0.75em}){9-9} \cmidrule(lr{0.75em}){10-10} \cmidrule(lr{0.75em}){11-11}
\cmidrule(lr{0.75em}){12-12} \cmidrule(lr{0.75em}){13-13}
Sem Poses &0.00 &11.37 &91.91 &\cellcolor{yellow!30}4.02 &-- &-- &-- &-- &-- &-- &-- &-- \\
Sem Voxels &-- &-- &-- &-- &20.49 &0.00 &100.00 &0.00 &13.03 &0.00 &100.00 &0.00 \\
Spherical Sem Fronts &-- &-- &-- &-- &20.49 &18.00 &82.35 &\cellcolor{orange!30}0.40 &13.03 &13.33 &87.02 &\cellcolor{orange!30}0.41 \\
Unidirectional Sem Fronts &-- &-- &-- &-- &20.49 &16.12 &85.93 &\cellcolor{yellow!30}3.15 &13.03 &11.58 &89.54 &\cellcolor{yellow!30}2.07 \\
\mapname \;(Ours)&0.00 &36.59 &75.37 &\cellcolor{green!25}\textbf{16.27} &20.49 &22.94 &81.15 &\cellcolor{green!25}\textbf{7.08} &13.03 &14.32 &88.69 &\cellcolor{green!25}\textbf{4.56} \\
\bottomrule
\end{tabular}}
\end{table*}

\noindent\textbf{Search Cut Volume Recall Metric:}
Our proposed metric, shown in Figure~\ref{fig:metrics}, measures how accurately and efficiently a mapping system cuts search volume. The intuitive definition is to compute total unmapped volume $vol_{unmapped}$ and subtract the search volume from it, however to avoid punishing true positives, we define search cut volume (SCV) as:
\begin{equation}
    SCV = 1 - \frac{FP_{unmapped}}{vol_{unmapped}}, \quad SCV \in [0, 1]
\end{equation}
To temper the metric against incorrectly cutting down volume, we compute Recall in the unmapped region and multiply it with SCV yielding the \textbf{Search Cut Volume Recall (SCVR)} metric:
\begin{equation}
    SCVR = SCV * \frac{TP_{unmapped}}{FN_{unmapped} + TP_{unmapped}}, \quad SCVR \in [0, 1]
\end{equation}
The \textbf{SCVR} metric is robust to both naive cases (1) not constraining the search volume, or (2) constraining it to 0 volume, yielding 0 for both. For an aggregate, we compute the area under the SCVR-time curve, stopping time for each class when 50\% of it has entered the mapped region. To further assess the robustness of \mapname, we vary depth sensing range at 0m, 10m, and 20m.

\begin{figure*}[!htb]
\centering
\includegraphics[width=0.9\textwidth]{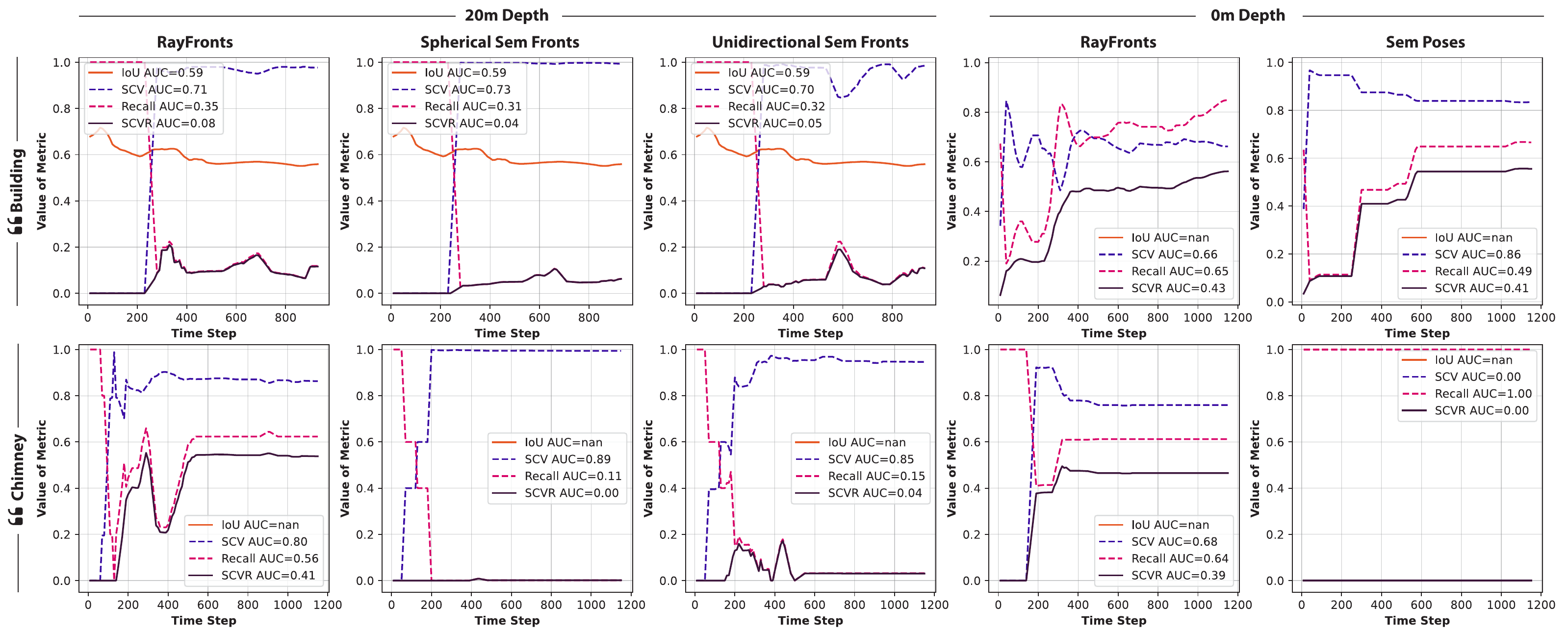} 
\caption{\textbf{\mapname{} consistently surpasses baselines for online semantic mapping}. Two query scenarios are shown: (1) querying for a prominent object (i.e Building) that enters depth range, and (2) a distant object (i.e Chimney) that remains beyond range. Through unified dense voxel mapping, and beyond-range semantic ray frontiers, \mapname{} sets the upper-bound in both scenarios.}
\label{fig:baselines}
\end{figure*}

\subsection{Offline 3D Open-Vocabulary Semantic Segmentation}
\label{sec:experiments-3d-semantic-seg}

\noindent\textbf{Datasets:} We follow prior work \cite{conceptfusion, conceptgraphs, hovsg-werby} and evaluate on Replica (\texttt{office[0-4]}, \texttt{room[0-2]}) and ScanNet (\texttt{scene[0011,0050,0231,0378,0518]}). In line with previous protocols, we report results while ignoring background classes (``\textit{floor}", ``\textit{wall}", ``\textit{ceiling}", ``\textit{door}", ``\textit{window}"). However, we additionally evaluate across all classes to demonstrate our ability to handle background seamlessly. Moreover, to showcase \mapname{}'s effectiveness in outdoor, unstructured, ``in-the-wild" environments, we further evaluate on the TartanAirV2 \cite{wang2020tartanair} scenes referenced in \ref{sec:experiments-online-mapping} excluding methods that cannot function outdoors.

\noindent\textbf{Baselines:} We compare our method with two categories of approaches: (1) vision-language representations that create 3D semantic maps, namely,  ConceptFusion \cite{conceptfusion}, ConceptGraphs \cite{conceptgraphs}, and HOV-SG\cite{hovsg-werby}; and (2) zero-shot semantic segmentation encoders, namely, NACLIP\cite{naclip} and Trident\cite{shi2024harnessing}. We extend the latter encoder-based methods to 3D using the same projection and fusion method as our system.

\noindent\textbf{Evaluation Protocol:} We follow standard open-vocabulary semantic segmentation evaluation protocols. We generate 3D segmentations by running HOV-SG and ConceptGraph code ensuring an accurate representation of their scene graph method. For all others, we generate segmentations by computing the cosine similarity between the embedded feature and the class-name text embedding, making a voxel prediction if its softmax probability exceeds $0.1$. 
We encode class names using each method's specified templates. For our approach, we follow NACLIP and use 80 templates \cite{naclip}, with a prompt denoising \cite{zhou2022extract} threhsold of $0.5$ to suppress irrelevant classes. We also apply k-NN matching (\texttt{k=5}) following HOV-SG \cite{hovsg-werby} protocol, assigning each GT voxel the majority label. All baselines use the ViT-L model architecture for consistency.
We resize images to 480x640, apply a frame skip of 10, 5cm voxels for Replica and ScanNet and 1m voxels for TartanAir.

\begin{table*}[!t]
\centering
\caption{Offline 3D Semantic Segmentation Benchmarking on Indoor Datasets.}\label{tab:semseg}
\resizebox{\textwidth}{!}{
\begin{tabular}{lcccccccccccc}
\toprule
& \multicolumn{6}{c}{\textbf{Replica}~\cite{straub2019replica}} & \multicolumn{6}{c}{\textbf{ScanNet}~\cite{dai2017scannet}} \\ \cmidrule{2-7} \cmidrule(lr{0.75em}){8-13}

& \multicolumn{3}{c}{\textbf{Without Background}} & \multicolumn{3}{c}{\textbf{With Background}} & \multicolumn{3}{c}{\textbf{Without Background}} & \multicolumn{3}{c}{\textbf{With Background}} \\ \cmidrule{2-4} \cmidrule(lr{0.75em}){5-7} \cmidrule(lr{0.75em}){8-10}
\cmidrule(lr{0.75em}){11-13}

{\textbf{Methods}} & \multicolumn{1}{l}{mIoU(\%)} & \multicolumn{1}{l}{f-mIoU(\%)} & \multicolumn{1}{l}{Acc(\%)} & \multicolumn{1}{l}{mIoU(\%)} & \multicolumn{1}{l}{f-mIoU(\%)} & \multicolumn{1}{l}{Acc(\%)} & \multicolumn{1}{l}{mIoU(\%)} & \multicolumn{1}{l}{f-mIoU(\%)} & \multicolumn{1}{l}{Acc(\%)} & \multicolumn{1}{l}{mIoU(\%)} & \multicolumn{1}{l}{f-mIoU(\%)} & \multicolumn{1}{l}{Acc(\%)} \\ \cmidrule{1-1} \cmidrule(lr{0.75em}){2-2} \cmidrule(lr{0.75em}){3-3} \cmidrule(lr{0.75em}){4-4} \cmidrule(lr{0.75em}){5-5} \cmidrule(lr{0.75em}){6-6} \cmidrule(lr{0.75em}){7-7} \cmidrule(lr{0.75em}){8-8} \cmidrule(lr{0.75em}){9-9} \cmidrule(lr{0.75em}){10-10} \cmidrule(lr{0.75em}){11-11}
\cmidrule(lr{0.75em}){12-12} \cmidrule(lr{0.75em}){13-13}

ConceptFusion \cite{conceptfusion} & \cellcolor{orange!30}21.07 & 31.51 & 35.65 & \cellcolor{orange!30}20.38 & \cellcolor{orange!30}35.75 & \cellcolor{orange!30}41.58 & 21.76 & 26.71 & 34.13 & 18.57 & 23.06 & 28.77 \\
ConceptGraphs \cite{conceptgraphs} & 11.63 & 16.61 & 19.80 & 11.72 & 21.35 & 28.28 & 21.62 & 24.32 & 31.04 & 20.83 & 23.61 & 35.80 \\
HOV-SG \cite{hovsg-werby} & 16.93 & 31.45 & 34.74 & 19.29 & 30.64 & 35.17 & 26.79 & 36.05 & 44.17 & \cellcolor{orange!30}23.48 & \cellcolor{yellow!30}28.92 & \cellcolor{yellow!30}38.52 \\
NACLIP-3D \cite{naclip} & 20.37 & \cellcolor{orange!30}35.08 & \cellcolor{orange!30}47.47 & 15.30 & 16.98 & 26.23 & \cellcolor{yellow!30}31.66 & \cellcolor{yellow!30}39.03 & \cellcolor{yellow!30}51.65 & 22.32 & 24.32 & 33.46 \\
Trident-3D \cite{shi2024harnessing} & \cellcolor{yellow!30}21.30 & \cellcolor{yellow!30}43.34 & \cellcolor{yellow!30}54.79 & \cellcolor{yellow!30}20.63 & \cellcolor{yellow!30}38.53 & \cellcolor{yellow!30}50.31 & \cellcolor{orange!30}29.97 & \cellcolor{orange!30}37.62 & \cellcolor{orange!30}51.06 & \cellcolor{yellow!30}24.80 & \cellcolor{orange!30}27.77 & \cellcolor{orange!30}38.43 \\
\mapname \;(Ours) & \cellcolor{green!25}\textbf{39.37} & \cellcolor{green!25}\textbf{62.03} & \cellcolor{green!25}\textbf{68.80} & \cellcolor{green!25}\textbf{27.73} & \cellcolor{green!25}\textbf{43.37} & \cellcolor{green!25}\textbf{54.45} & \cellcolor{green!25}\textbf{41.29} & \cellcolor{green!25}\textbf{46.42} & \cellcolor{green!20}\textbf{56.76} & \cellcolor{green!25}\textbf{32.29} & \cellcolor{green!25}\textbf{39.04} & \cellcolor{green!25}\textbf{49.15} \\ \hline
\end{tabular}}
\end{table*}

\begin{table}[!htp]\centering
\caption{Offline 3D Semantic Segmentation Benchmarking on an Outdoor Dataset (TartanAirV2~\cite{wang2020tartanair}).}\label{tab:tv2}
\scriptsize
\begin{tabular}{lccc}\toprule
\textbf{Methods} & mIoU (\%) & f-mIoU (\%) & Acc (\%) \\
\cmidrule{1-1} \cmidrule(lr{0.75em}){2-2} \cmidrule(lr{0.75em}){3-3} \cmidrule(lr{0.75em}){4-4}
ConceptFusion \cite{conceptfusion} & 5.84 & 32.78 & 39.76 \\
NACLIP-3D \cite{naclip} & \cellcolor{orange!30}9.66 & \cellcolor{orange!30}40.82 & \cellcolor{orange!30}54.10 \\
Trident-3D \cite{shi2024harnessing}& \cellcolor{yellow!30}9.86 & \cellcolor{green!25}\textbf{43.56} & \cellcolor{yellow!30}55.34 \\
\mapname{} (Ours) & \cellcolor{green!25}\textbf{13.22} & \cellcolor{yellow!30}43.43 & \cellcolor{green!25}\textbf{57.26} \\
\bottomrule
\end{tabular}
\end{table}

\section{Results \& Discussion}
\label{sec:results}
\subsection{Online Semantic Mapping}
\cref{tab:mapping} summarizes online performance of the five methods in their respective operating ranges. We observe that \mapname{} excels and is the upper bound across depth ranges. \textbf{Sem Poses} fails to capture any fine-grained reconstructions scoring 0 mIoU-AUC, while \textbf{Sem Voxels} fails to provide any information about the unmapped region scoring 0 SCVR-AUC. At 0 depth range, \mapname{} attaches dense semantic rays at each pose as opposed to the global encoding scheme employed by \textbf{Sem Poses}. This allows us to encode non-prominent objects seamlessly and results in a $\sim4\times$  SCVR-AUC than \textbf{Sem Poses}. This observation is illustrated in \cref{fig:baselines} where for a simple prominent object such as ``building", both \textbf{Sem Poses} and \mapname{} perform similarly. However, for a more distant object like ``chimney", \textbf{Sem Poses} fails to capture its semantics entirely. At higher depth ranges, we observe that \mapname{} consistently outperforms semantic frontier baselines at $\sim2.2\times$ the SCVR-AUC. We attribute this to (1) less semantic collisions as distinct objects are unlikely to be fused in the same ray unlike semantic frontiers which can have many collisions, (2) better preservation of the angle at which the semantic entity was observed from, and (3) allowing each frontier to have multiple rays attached, increasing the density of beyond-range semantics. \mapname{} is superior to all baselines across depth ranges \textbf{empowering both fine-grained localization and beyond-range guidance}.

\subsection{Offline 3D Semantic Segmentation}

\cref{tab:semseg} provides a detailed comparison of the performance between our framework and other zero-shot approaches, outlined in \cref{sec:experiments-3d-semantic-seg}. \mapname{} consistently outperforms the baselines in mIoU, \textbf{and achieves SOTA performance} beating the next best baselines by +18.07\% and +9.63\% mIoU on Replica and Scannet, respectively, excluding background. \mapname{} is also able to handle background seamlessly with its single-forward pass approach while segment-and-encode approaches fall short.

For outdoor in-the-wild performance on TartanAirV2, \cref{tab:tv2} shows that \mapname{} exceeds the performance of the baselines by 3.36\% mIoU. While Trident-3D serves as a close second to our approach and achieves a slightly higher f-mIoU on TartanAirV2 by a marginal 0.13\%, it does so at the cost of integrating multiple foundational models into their pipeline, which significantly reduces efficiency—an essential factor for online semantic mapping.

\subsection{Encoder \& Mapping Throughput Analysis}
\begin{figure}[t]
    \centering
    \includegraphics[width=0.9\linewidth]{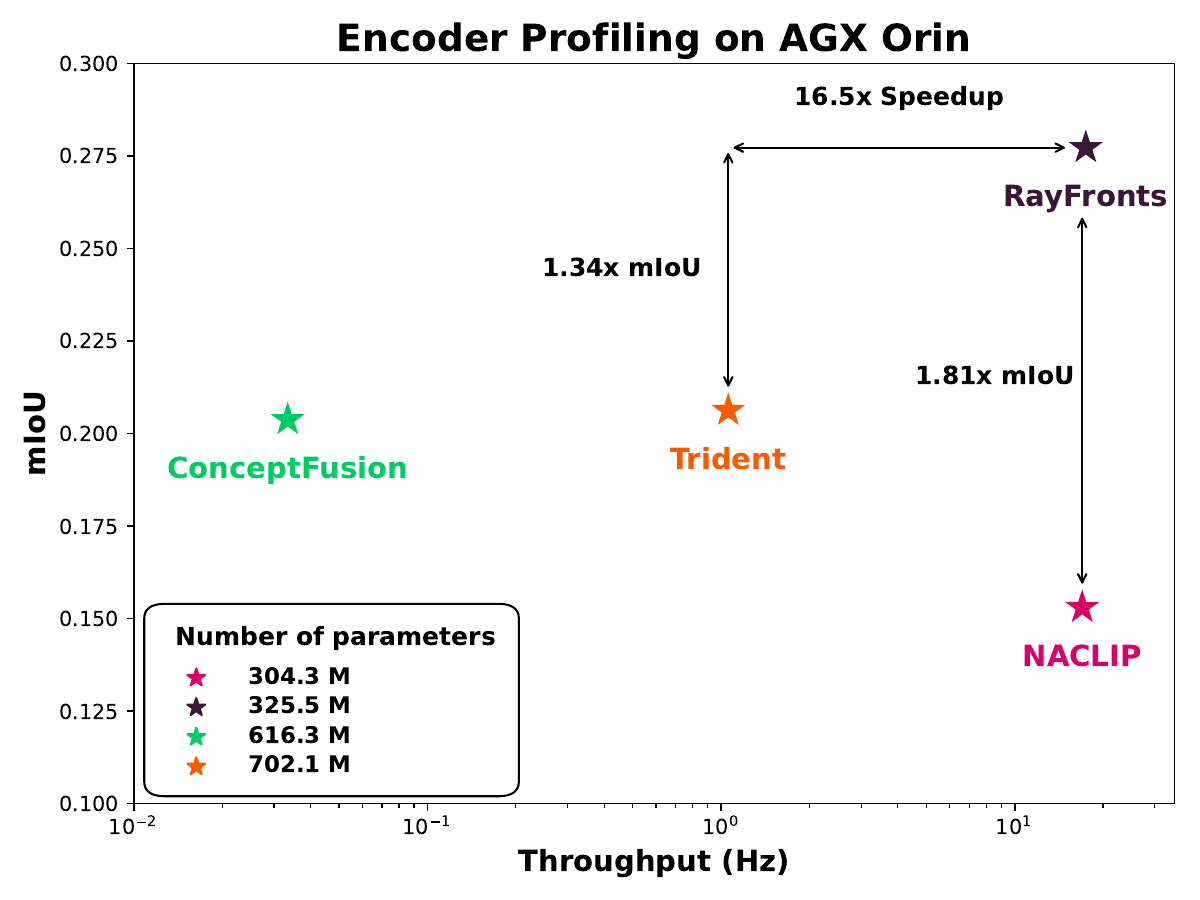}
    \caption{\textbf{\mapname{} provides state-of-the-art mIoU \& 17.5 Hz throughput on an AGX Orin.} It surpasses Trident with 1.34x higher mIoU and a 16.5x speedup, while achieving 1.81x higher mIoU than NACLIP, which operates at a similar throughput.}
    \label{fig:orin}
    \vspace{-0.5cm}
\end{figure}

To assess deployability, we run \mapname{} on an NVIDIA Jetson AGX Orin and perform a quantitative comparison of image encoder throughput shown in \cref{fig:orin}. Our mapping system achieves SOTA performance in 3D open-set semantic segmentation with 1.34x the mIoU of Trident while being 16.5x faster, running at 17.5 Hz, and with only 46\% of the parameters. While NACLIP has similar throughput, we surpass it by a significant 1.81x in mIoU. ConceptFusion's 0.03 Hz throughput makes it impractical for real-time use. 

Furthermore, we test the end-to-end throughput of \mapname{} on a real-world outdoor scene using pre-recorded data from a mobile ground robot. We use a resolution of 224x224, 30cm voxel size, and the base encoder model, while compressing features to top 100 PCA components (retaining $\sim 80\%$ variance), and disabling ray-tracing. \textbf{\mapname{} runs real-time at 8.84 Hz on Orin AGX.}

\subsection{Qualitative Real-World Study} 
To evaluate \mapname{} in unbounded, open-world settings, we record a run through an unstructured fire training facility with a Zed-X camera. As shown in \cref{fig:fig1}, \mapname{} accurately reconstructs fine-grained details (e.g., ``road cracks") while detecting far-range objects (e.g., ``water tower"), demonstrating that \mapname{} \textbf{empowers robots within and beyond depth-sensing limitations in open-world environments}.

\section{Conclusion}
\label{sec:conclusion}

We present \mapname{}, a real-time semantic mapping system for multi-modal open-set scene understanding for both within- and beyond-range mapping. Our key insight, \textit{semantic ray frontiers}, enables open-set queries about observations beyond depth mapping by associating beyond-depth ray features with the map's frontiers. This allows \mapname{} to significantly reduce search volumes while retaining fine-grained within-range scene understanding. \mapname{} improves open-set image encoding with an efficient language-aligned encoder, and introduces a new planner-agnostic metric for open-world search. We achieve state-of-the-art results in 3D open-set semantic segmentation, strong performance in online mapping, and efficient encoder throughput. Our future work aims to include instance differentiation in \mapname{} and planning integration to facilitate online exploration.

\ifarxiv
\section*{Limitations}
\label{sec:limitations}
While \mapname{} is the upper-bound of the online mapping baselines in correct search volume reduction, it also has the highest memory consumption. However, \mapname{} can be tuned down by reducing ray angle bins down until  a single bin (becoming ``semantic frontiers"), or reducing depth range down until 0, giving the flexibility to achieve the best trade-off for an application.

\fi

\section*{Acknowledgments}
\ifarxiv
This work was supported by Defense Science and Technology Agency (DSTA) Contract \#DST000EC124000205, King Abdulaziz University, and National Institute on Disability, Independent Living, and Rehabilitation Research (NIDILRR) Grant \#90IFDV0042.
We thank Krishna Murthy Jatavallabhula, Jacob Yeung, Sam Triest and Ayush Jain for insightful discussions and feedback, and Katerina Nikiforova for early encoder exploration.
\fi
\ifiros
We thank Krishna M. Jatavallabhula, Jacob Yeung, Sam Triest and Ayush Jain for insightful discussions and feedback, and Katerina Nikiforova for early encoder exploration.
\fi

\ifarxiv

\section*{Appendix}

\setcounter{section}{0}
\setcounter{equation}{0}
\setcounter{figure}{0}
\setcounter{table}{0}

\renewcommand{\thesection}{A\arabic{section}}
\renewcommand{\thesubsection}{A\arabic{subsection}}
\renewcommand{\thefigure}{A.\arabic{figure}}
\renewcommand{\thetable}{A.\arabic{table}}
\renewcommand{\theequation}{A.\arabic{equation}}

\section{Contribution Statement}

\textbf{Omar Alama} led the research and conceptual design, developed the mapping codebase and online/offline evaluation scripts, wrote major sections of the paper, and created figures, online mapping tables, and videos.

\textbf{Avigyan Bhattacharya} developed the RayFronts encoder code, implemented the ScanNet data loader, and ported Trident for evaluation. Conducted extensive evaluations of multiple offline mapping baselines and wrote the corresponding sections and appendix figures.

\textbf{Haoyang He} ported ConceptFusion and NACLIP for evaluation, performed extensive evaluations of multiple offline mapping baselines, and conducted throughput analyses on the ORIN AGX. Also contributed to writing the corresponding sections.

\textbf{Seungchan Kim} engaged in early discussions, developed the TartanAirV2 data loader, and played a key role in refining the paper writing.

\textbf{Yuheng Qiu} engaged in early discussions, explored datasets for evaluation, and provided writing feedback.

\textbf{Wenshan Wang} participated in early discussions and provided feedback on research direction and paper writing.

\textbf{Cherie Ho} deployed RayFronts on the Orin AGX, assisted in collecting test data across various robot platforms, and contributed extensively to paper writing, coordination, and the design of figures and video.

\textbf{Nikhil Keetha} was heavily involved in brainstorming, discussions, and the conceptual design of the mapping, encoding, and evaluation frameworks. Contributed significantly to paper writing, as well as figures and video design.

\textbf{Sebastian Scherer} shaped the research area, engaged in discussions and brainstorming, and provided valuable feedback on writing, figures, and video.

\section{Online Semantic Mapping Visualizations}
\cref{fig:baselines_illustrations} shows illustrations of the different baselines mentioned in \cref{sec:experiments-online-mapping}. The top left part of the figure highlights how \mapname{} can avoid semantic feature collisions (the case where different semantic features are forced to be fused together) by utilizing multiple rays to describe the different semantic entities. Whereas semantic frontier approaches (irrespective of the unidirectional/spherical search volume method) can fail when semantically different entities are observed through the same frontier. The top right part of the figure emphasizes where global encoding approaches like Sem Poses can fail to capture non-prominent objects in the presence of a large centered entity. The illustration provides further explanation for Sem Poses's inability to capture chimneys in the AbandonedCableDay scene as shown in \cref{fig:baselines,fig:online_eval_vis}. Finally, the bottom row illustrates how each baseline computes its search volume. Spherical Sem Fronts can fail to capture a distant object, with increasing radius cubically increasing search volume. Unidirectional Sem Fronts is highly sensitive to the mapped region topology since it uses it to infer the semantic ray direction, and in the illustrated case, it fails. Sem Poses fails to utilize depth information to push the ray further onto the mapped region boundary for better localization. In contrast with all baselines, \mapname{} is able to accurately determine the direction of the ray and limit the search volume efficiently, utilizing depth information if available. 

\cref{fig:online_eval_vis} visualizes the two query scenarios shown in \cref{fig:baselines} with ground truth generated at an 80m cutoff (Highest value that fits in our memory) for further clarity. The top block shows search volumes for building at a particular time step. At 20m depth sensing range, it is clear that \mapname{} achieves the best search volume, having 1.35 $\times$ higher SCVR than Unidirectional Sem Fronts. Spherical Sem Fronts struggles to cast a search volume that encompasses big objects, whereas Unidirectional Sem Fronts has some erroneously inferred ray directions. At 0m range, both Sem Poses and \mapname{} perform similarly. Furthermore, the bottom block shows the search volume of distant non-prominent objects that never come into the depth sensing range. At 0m range, Sem Poses fails to capture the semantics and fails to reduce search volume, yielding an SCVR of 0, whereas \mapname{} provides meaningful areas to explore.

\begin{figure*}[ht]
\centering
\includegraphics[width=0.9\textwidth]{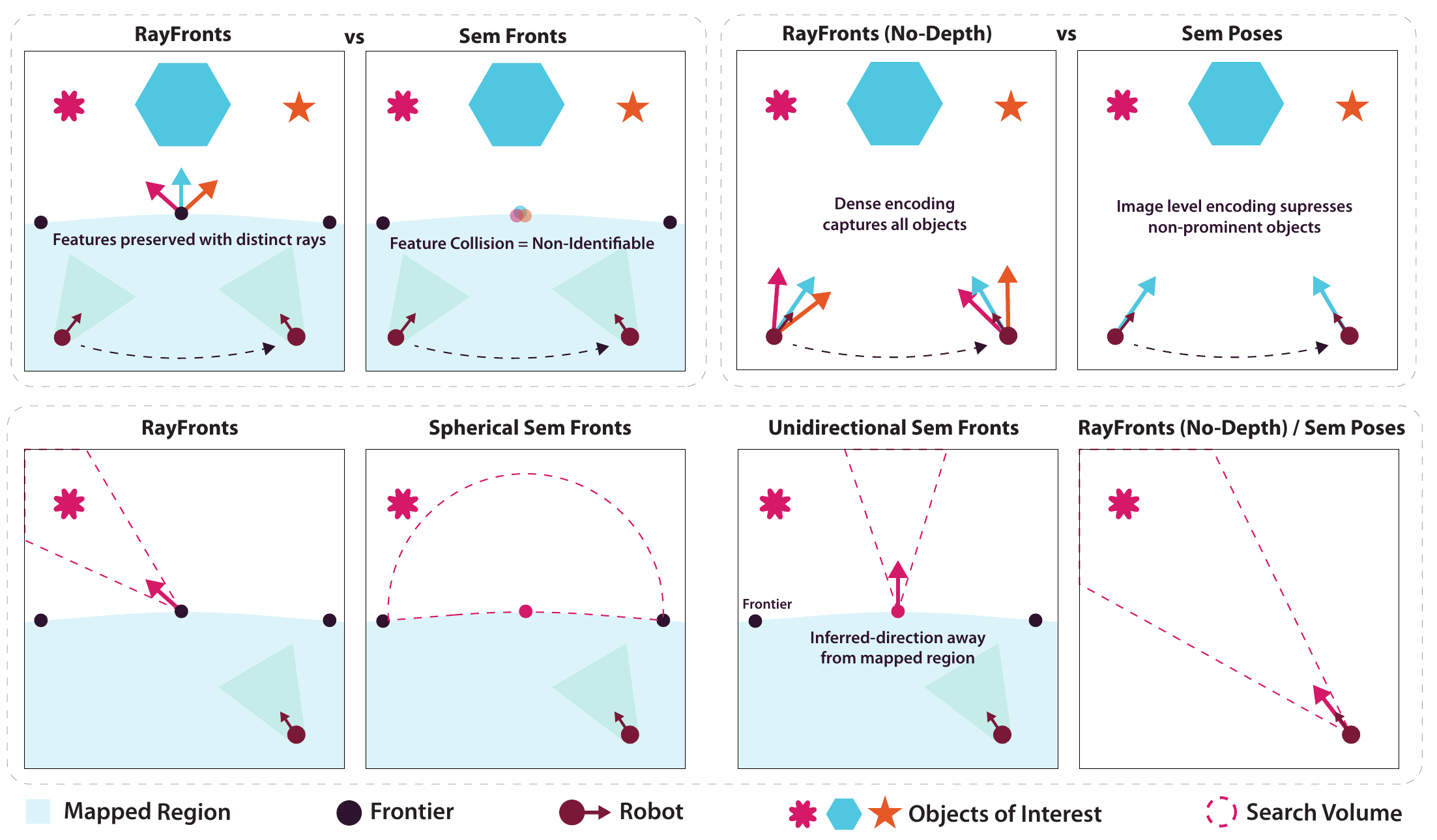} 
\caption{Top left shows how \mapname{} is able to avoid feature collisions through the use of multiple rays that capture distinct semantics observed through the same frontier, where semantic frontier approaches \cite{chen2023train, yokoyama2024vlfm} fail. The top right illustrates that even with no depth information, \mapname{} dense language-aligned encoding can allow it to capture non-prominent semantics where semantic pose approaches \cite{thomas2024embedding} fail. The bottom row highlights that \mapname{} is the upper bound in accurately reducing search volume.}
\label{fig:baselines_illustrations}
\end{figure*}

\begin{figure*}[ht]
\centering
\includegraphics[width=0.88\textwidth]{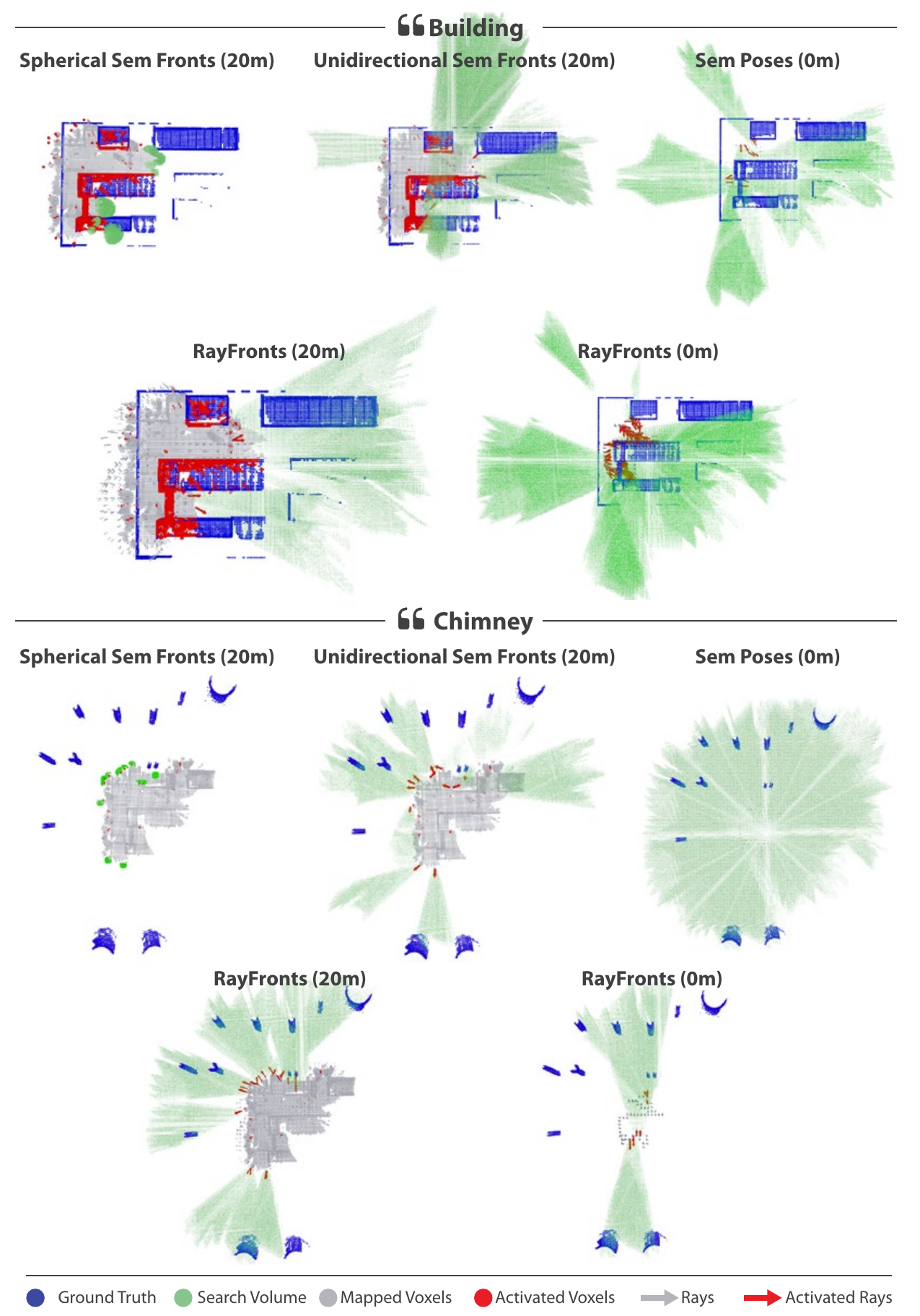} 
\caption{Two query scenarios are shown with GT generated at 80m as opposed to 50m cutoff for more clarity: (1) querying for a prominent object (i.e, Building) that enters depth range, and (2) a distant object (i.e, Chimney) that remains beyond range. Through unified dense voxel mapping and beyond-range semantic ray frontiers, \mapname{} sets the upper bound in both scenarios.}
\label{fig:online_eval_vis}
\end{figure*}

\section{Offline Semantic Mapping Visualizations}
\cref{fig:offline_semseg} showcases open-vocabulary semantic segmentation samples across different datasets, while \cref{fig:open_vocab} highlights the open-vocabulary capabilities of \mapname{} by showing the segmentations of multiple long-tail classes.
\begin{figure*}[!htb]
\centering
\includegraphics[width=\textwidth]{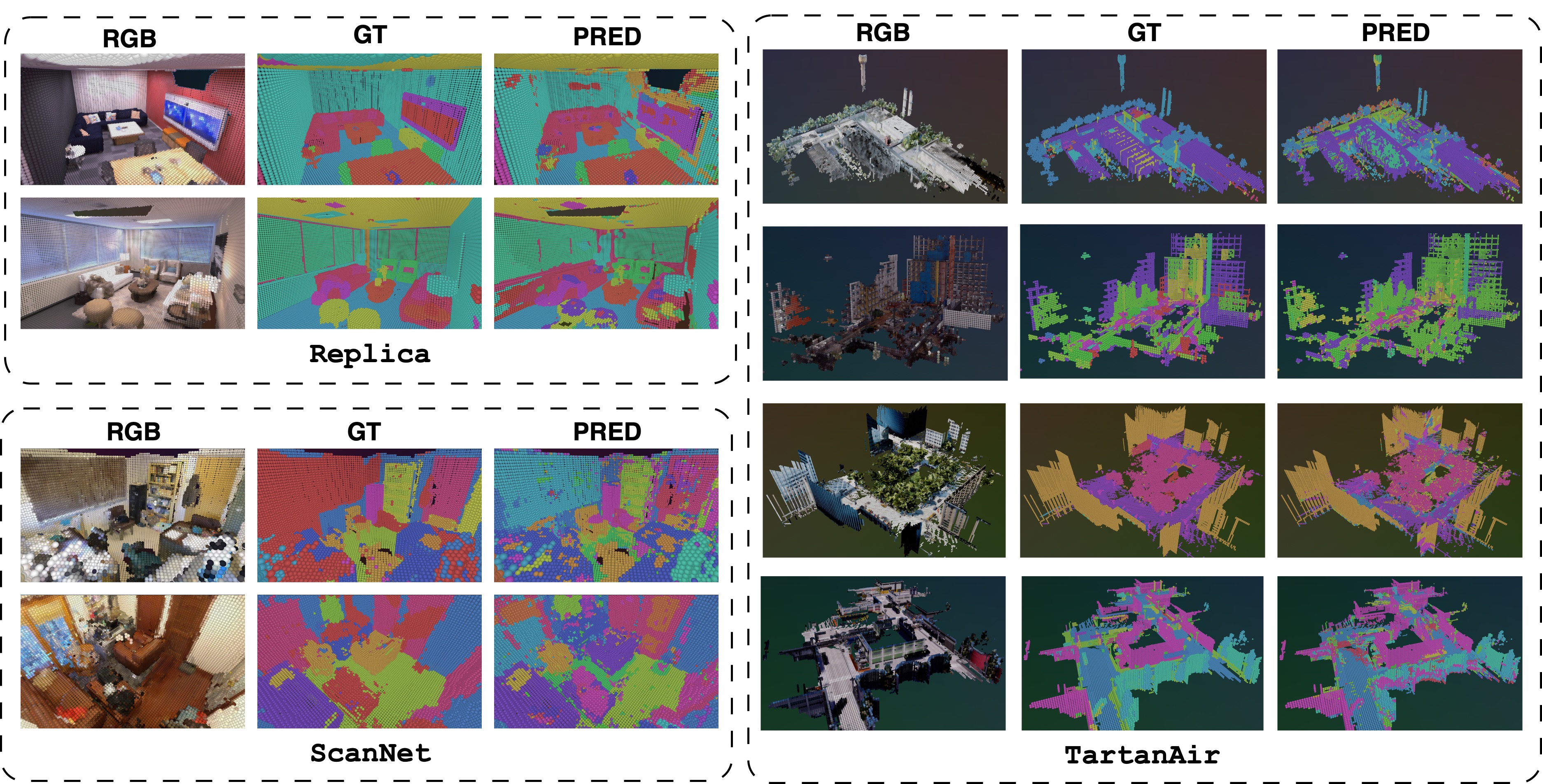} 
\caption{Sample visualizations of offline semantic mapping generated by \mapname{} for scenes from Replica \cite{straub2019replica} (\texttt{room0} and \texttt{office2}), ScanNet \cite{dai2017scannet}\texttt{(scene0050} and \texttt{scene00378}), and the chosen four scenes from TartanAir \cite{wang2020tartanair}. ``RGB", ``GT" and ``PRED" refer to the RGB scene reconstruction, Ground Truth semantics, and semantic segmentation prediction by \mapname{}, respectively, for each corresponding scene. \mapname{} achieves SOTA mIoU for 3D open-vocabulary semantic segmentation.}
\label{fig:offline_semseg}
\end{figure*}

\begin{figure*}[!htb]
\centering
\includegraphics[width=0.99\textwidth]{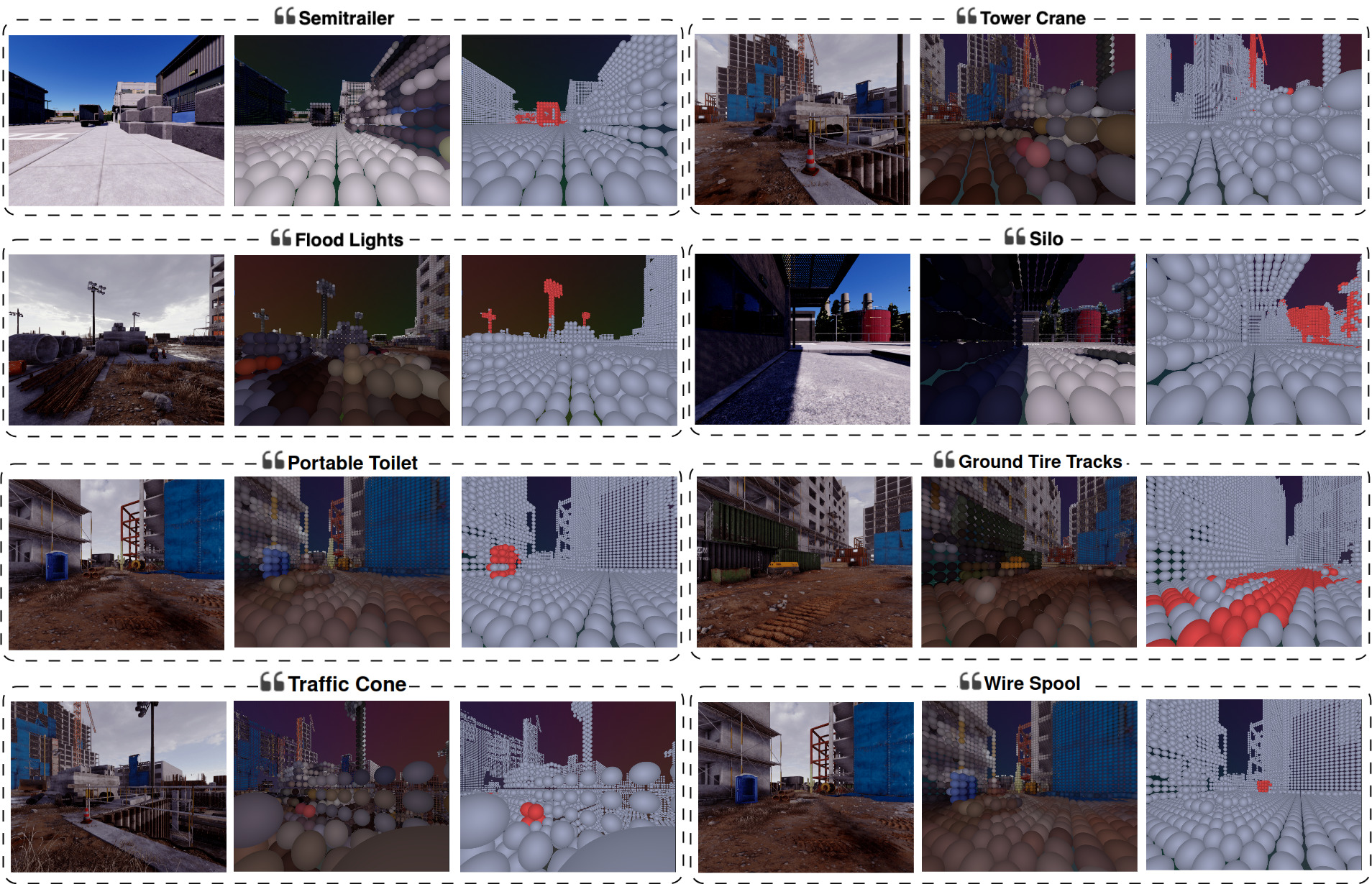} 
\caption{Examples of long-tail classes segmented by \mapname{} across outdoor scenes from TartanAir \cite{wang2020tartanair}. We set the voxel size to 0.5 (50cm) for the visualizations. For each set, we present the RGB image, the corresponding 3D reconstructed view, and the classified voxels left to right respectively. \mapname{} effectively segments long-tail concepts.}
\label{fig:open_vocab}
\end{figure*}

\section{RayFronts Hyperparameters}
\cref{tab:hyper_desc} lists all the \mapname{} hyperparameters and their descriptions, \cref{tab:hyper_online} lists the hyperparameter values used for the online mapping evaluation, \cref{tab:hyper_offline} lists the hyperparameter values used for the offline mapping evaluation, and \cref{tab:hyper_throughput} lists the hyperparameter values used for the throughput analysis.

\begin{table*}[!t]
\caption{\mapname{} Hyperparameter Descriptions.}\label{tab:hyper_desc}
\centering
\resizebox{0.9\textwidth}{!}{
\resizebox{\textwidth}{!}{
\begin{tabular}{lll}
\hline
\multicolumn{1}{c}{parameter}               & \multicolumn{1}{c}{description}                                                  \\ \hline
$backbone$                                  & Backbone model used.                                                             \\
$resolution$                                & Input RGB and depth resolution.                                                  \\
$gauss\_std$ ($\sigma$)                     & Standard deviation of the Gaussian kernel attention for the encoder.             \\
$vox\_size$ ($\alpha$)                      & Voxel size in meters.                                                            \\
$fronti\_neighborhood\_r$                   & Radius of  the neighborhood to look at for computing frontiers.                  \\
$fronti\_min\_unobserved $($min_{unobsrv}$) & Min \# of unobserved cells in the cell neighborhood to be considered a frontier. \\
$fronti\_min\_occupied$ ($min_{occ}$)       & Min \# of occupied cells in the cell neighborhood to be considered a frontier.   \\
$fronti\_min\_empty$ ($min_{free}$)         & Min \# of free cells in the cell neighborhood to be considered a frontier.       \\
$fronti\_subsampling$                       & Subsampling factor of the frontier grid. $\beta= \alpha*fronti\_subsampling$.    \\
$fronti\_subsampling\_min\_fronti$          & \# of frontiers to lie in the coarser grid cell to be considered a frontier.     \\
$ray\_erosion$                              & Half size of the window to use when eroding the out-of-range mask $M_t$.         \\
$ray\_tracing$                              & Enable or disable ray tracing when propagating rays.                             \\
$angle\_bin\_size$ ($\psi$)                 & Angle bin size used to discretize and aggregate rays within a frontier.          \\
$max\_occ\_cnt$                             & Log-odds upper limit for occupancy.                                              \\
$max\_empty\_cnt$                           & Log-odds lower limit for occupancy.                                              \\
$occ\_observ\_weight$                       & How much to increment the log odds buffer with each occupied observation.        \\
$occ\_thickness$                            & Thickness of a projected occupied surface.                                       \\
$occ\_pruning\_tolerance$                   & Tolerance of log-odds value to be merged in a super voxel in the VDB map.        \\
$max\_dirs\_per\_frame$                     & Max number of rays to cast per frame. (Uniformly samples to enforce).            \\
$max\_pts\_per\_frame$                      & Max number of occupied points to unproject. (Uniformly samples to enforce).      \\
$max\_empty\_pts\_per\_frame$               & Max number of empty points to unproject. (Uniformly samples to enforce).         \\
$vox\_accum\_period$                        & How many frames should accumulate before aggregating voxels.                     \\
$ray\_accum\_period$                        & How many frames should accumulate before casting and aggregating rays.           \\
$ray\_accum\_phase$                         & Ray accumulation delay to offset it from voxel accumulation.                     \\
$stored\_feat\_dim$                         & Dimension of map features. If less than encoder output, PCA is used to compress.                    \\
$sem\_pruning\_period$                         & How often to prune semantic voxels using the occupancy map.        \\
$occ\_pruning\_period$                         & How often to prune the occupancy map (i.e merge large consistent areas).        \\
$prompt\_denoising\_thresh$                         & Threshold for prompt denoising when classifying voxels/rays).   \\
$prediction\_thresh$                         & If the softmax value was lower than this threshold, no prediction will be made.   \\
$searchvol\_thresh$                         & Threshold to select the intersection of multiple search volumes. 

\end{tabular}}}
\end{table*}
\begin{table}[!ht]
\caption{\mapname{} Online Evaluation Hyperparameters.}\label{tab:hyper_online}
\centering
\begin{tabular}{ll}
\hline
\multicolumn{1}{c}{parameter}              & value                  \\ \hline
$backbone$                                 & radio\_v2.5-l / SIGLIP \\
$resolution$                               & 640x640                \\
$gauss\_std$ ($\sigma$)                    & 7.0                    \\
$vox\_size$ ($\alpha$)                     & 1.0                    \\
$fronti\_neighborhood\_r$                  & 1                      \\
$fronti\_min\_unobserved$($min_{unobsrv}$) & 9                      \\
$fronti\_min\_occupied$ ($min_{occ}$)      & 0                      \\
$fronti\_min\_empty$ ($min_{free}$)        & 4                      \\
$fronti\_subsampling$                      & 4                      \\
$fronti\_subsampling\_min\_fronti$         & 5                      \\
$ray\_erosion$                             & 32                     \\
$ray\_tracing$                             & True                   \\
$angle\_bin\_size$ ($\psi$)                & 30\textdegree          \\
$max\_occ\_cnt$                            & 100                    \\
$max\_empty\_cnt$                          & -10                    \\
$occ\_observ\_weight$                      & 100                    \\
$occ\_thickness$                           & 2                      \\
$occ\_pruning\_tolerance$                  & 2                      \\
$max\_dirs\_per\_frame$                    & 10000                  \\
$max\_pts\_per\_frame$                     & $+\infty$                \\
$max\_empty\_pts\_per\_frame$              & $+\infty$               \\
$stored\_feat\_dim$                         & 768 \\
$prompt\_denoising\_thresh$                         & 0.5  \\
$prediction\_thresh$                         & 0.1   \\
$searchvol\_thresh$                         & 0.05
\end{tabular}
\end{table}

\begin{table}[!ht]
\caption{\mapname{} Offline Evaluation Hyperparameters.}\label{tab:hyper_offline}
\centering
\begin{tabular}{ll}
\hline
\multicolumn{1}{c}{parameter}              & value                  \\ \hline
$backbone$                                 & radio\_v2.5-l / SIGLIP \\
$resolution$                               & 640x480                \\
$gauss\_std$ ($\sigma$)                    & 7.0                    \\
$vox\_size$ ($\alpha$)                     & 0.05, 1.0              \\
$stored\_feat\_dim$                         & 768 \\
$prompt\_denoising\_thresh$                         & 0.5  \\
$prediction\_thresh$                         & 0.1   
\end{tabular}
\end{table}

\begin{table}[]
\caption{\mapname{} Mapping Throughput Hyperparameters.}\label{tab:hyper_throughput}
\centering
\begin{tabular}{ll}
\hline
\multicolumn{1}{c}{parameter}              & value                  \\ \hline
$backbone$                                 & radio\_v2.5-b / SIGLIP \\
$resolution$                               & 224x224                \\
$gauss\_std$ ($\sigma$)                    & 7.0                    \\
$vox\_size$ ($\alpha$)                     & 0.3                    \\
$fronti\_neighborhood\_r$                  & 1                      \\
$fronti\_min\_unobserved$($min_{unobsrv}$) & 9                      \\
$fronti\_min\_occupied$ ($min_{occ}$)      & 0                      \\
$fronti\_min\_empty$ ($min_{free}$)        & 4                      \\
$fronti\_subsampling$                      & 4                      \\
$fronti\_subsampling\_min\_fronti$         & 5                      \\
$ray\_erosion$                             & 0                     \\
$ray\_tracing$                             & False                   \\
$angle\_bin\_size$ ($\psi$)                & 30\textdegree          \\
$max\_occ\_cnt$                            & 100                    \\
$max\_empty\_cnt$                          & -10                    \\
$occ\_observ\_weight$                      & 100                    \\
$occ\_thickness$                           & 1                      \\
$occ\_pruning\_tolerance$                  & 2                      \\
$max\_dirs\_per\_frame$                    & 1000                  \\
$max\_pts\_per\_frame$                     & 3000                \\
$max\_empty\_pts\_per\_frame$              & 10000               \\
$stored\_feat\_dim$                         & 100 \\
$vox\_accum\_period$                        & 8                     \\
$ray\_accum\_period$                        & 8           \\
$ray\_accum\_phase$                         & 4                     \\

$sem\_pruning\_period$                         & 32        \\
$occ\_pruning\_period$                         & 32  
\end{tabular}
\end{table}

\newpage
\newpage
\fi

\bibliographystyle{ieeetr}
\bibliography{root}

\end{document}